\documentclass[10pt,twocolumn,letterpaper]{article}

\usepackage{iccv}
\usepackage{times}
\usepackage{graphicx}
\usepackage{amsmath}
\usepackage{amssymb}
\usepackage{makeidx}
\usepackage{subfig}
\usepackage{float}
\usepackage{tabularx}
\usepackage{makecell}
\usepackage{multirow}
\usepackage{algorithm}
\usepackage{algpseudocode}
\usepackage[justification=centering]{caption}
\usepackage{nth}
\usepackage{balance}

\newcommand{\norm}[1]{\left\lVert#1\right\rVert}
\newcommand{\abs}[1]{|#1|}

\usepackage[pagebackref=true,breaklinks=true,letterpaper=true,colorlinks,bookmarks=false]{hyperref}

\iccvfinalcopy 

\def\iccvPaperID{6055} 

\ificcvfinal\fi
\begin{document}

\title{CroP: Color Constancy Benchmark Dataset Generator}

\author{Nikola Bani{\'{c}}\\{\tt\small nikola.banic@fer.hr} \and Karlo Ko{\v{s}}{\v{c}}evi{\'{c}}\\{\tt\small karlo.koscevic@fer.hr} \and Marko Suba{\v{s}}i{\'{c}}\\{\tt\small marko.subasic@fer.hr} \and Sven Lon{\v{c}}ari{\'{c}}\\{\tt\small sven.loncaric@fer.hr}\\
Image Processing Group\\
Faculty of Electrical Engineering and Computing\\
University of Zagreb, Unska 3, 10000 Zagreb, Croatia\\
}

\maketitle

\begin{abstract}
Implementing color constancy as a pre-processing step in contemporary digital cameras is of significant importance as it removes the influence of scene illumination on object colors. Several benchmark color constancy datasets have been created for the purpose of developing and testing new color constancy methods. However, they all have numerous drawbacks including a small number of images, erroneously extracted ground-truth illuminations, long histories of misuses, violations of their stated assumptions, etc. To overcome such and similar problems, in this paper a color constancy benchmark dataset generator is proposed. For a given camera sensor it enables generation of any number of realistic raw images taken in a subset of the real world, namely images of printed photographs. Datasets with such images share many positive features with other existing real-world datasets, while some of the negative features are completely eliminated. The generated images can be successfully used to train methods that afterward achieve high accuracy on real-world datasets. This opens the way for creating large enough datasets for advanced deep learning techniques. Experimental results are presented and discussed. The source code is available at \url{http://www.fer.unizg.hr/ipg/resources/color\_constancy/}.
\end{abstract}

\section{Introduction}
\label{sec:introduction}
Color constancy is the ability of the human vision system (HVS) to perceive the colors of the objects in the scene largely invariant to the color of the light source [25]. Most of the contemporary digital cameras have this ability implemented into their image pre-processing pipeline [40]. The task of computational color constancy is to estimate the scene illumination and then perform the chromatic adaptation in order to remove the influence of the illumination color on the colors of the objects in the scene. Three physical variables can describe the perceived color of objects in the image: 1)~spectral properties of the light source, 2)~spectral reflectance properties of the object surface, and 3)~spectral sensitivity of the camera sensor. Under the Lambertian assumption, the resulting image $\mathbf{f}$ formation model is
\begin{equation}
\label{eq:image}
    f_c(\mathbf{x}) = \int_\omega I(\lambda, \mathbf{x})R(\mathbf{x}, \lambda) \rho_c(\lambda)d\lambda
\end{equation}
where $f_c(\mathbf{x})$ is the value at the pixel location $\mathbf{x}$ for the $c$-th color channel, $I(\mathbf{x}, \lambda)$ is the spectral distribution of light source, $R(\mathbf{x}, \lambda)$ is the surface reflectance, and $\boldsymbol{\rho}(\lambda)$ is the camera sensor sensitivity for the $c$-th color channel. The value at pixel location $\mathbf{x}$ is obtained by integrating across the all wavelengths $\lambda$ of the light in the visible spectrum $\omega$. When estimating the illumination it is often assumed that it is uniform across the whole scene. With this, $\mathbf{x}$ can be disregarded and the observed light source $\mathbf{e}$ is calculated as
\begin{equation}
\label{eq:e}
    \mathbf{e} = \begin{pmatrix} e_R \\ e_G \\ e_B\end{pmatrix} = \int_\omega I(\lambda)\mathbf{\boldsymbol{\rho}}(\lambda)d\lambda.
\end{equation}
Since only pixel values $\mathbf{f}$ are known and both $I(\lambda)$ and $\boldsymbol{\rho}(\lambda)$ remain unknown, it is an ill-posed problem to calculate the illumination vector $\mathbf{e}$. Illumination estimation methods try solve this problem by introduction of new assumptions. On one side, there are methods that rely on low-level image statistics such as White-patch~\cite{land1977retinex,funt2010rehabilitation} and its improvements~\cite{banic2013using,banic2014color,banic2014improving}, Gray-world~\cite{buchsbaum1980spatial}, Shades-of-Gray~\cite{finlayson2004shades}, Gray-Edge (\nth{1} and \nth{2} order)~\cite{van2007edge}, using bright and dark colors~\cite{cheng2014illuminant}, exploiting the illumination color statistics perception~\cite{banic2019blue}, exploiting the expected illumination statistics~\cite{banic2018green}, using gray pixels~\cite{quian2019revisiting}. Appropriately, these methods can be found in the literature as statistics-based methods. They are fast, hardware-friendly, and easy to implement. On the other hand, there are learning-based methods, which use data to learn their parameter values and compute more precise estimations, but they also require significantly more computational power and parameter tuning. Learning-based method include gamut mapping~(pixel, edge, and intersection based)~\cite{finlayson2006gamut}, using high-level visual information~\cite{van2007using}, natural image statistics~\cite{gijsenij2007color}, Bayesian learning~\cite{gehler2008bayesian}, spatio-spectral learning~(maximum likelihood estimate, and with gen. prior)~\cite{chakrabarti2012color}, simplifying the illumination solution space~\cite{banic2015color,banic2015using,banic2015acolor}, using color/edge moments~\cite{finlayson2013corrected}, using regression trees with simple features from color distribution statistics~\cite{cheng2015effective}, performing various spatial localizations~\cite{barron2015convolutional,barron2017fast}, genetic algorithms and illumination restriction~\cite{koscevic2019color}, convolutional neural networks~\cite{bianco2015color,shi2016deep,hu2017fc4,qiu2018pilot}.

To compare the accuracy of these methods, several publicly available color constancy datasets have been created. While they significantly contributed to the advance of the illumination estimation, they have several drawbacks. The main one is that they contain relatively few images due to the significant amount of time required for determining the ground-truth illumination. This was shown to have an impact on the applicability of the deep learning techniques. Other common drawbacks include cases of incorrect ground-truth illumination data, significant noise amounts, violations of some important assumptions, etc. In the worst cases the whole datasets are being used completely wrong in the pure technical sense~\cite{banic2019past}, which may have led to many erroneous conclusions in the field of illumination estimation~\cite{finlayson2017curious}. In order to try to simultaneously deal with most of these problems, in this paper a color constancy dataset generator is proposed. It is confined only to simulation of taking images of printed photographs under projector illumination of specified colors, but in terms of illumination estimation the properties of the resulting images are shown to resemble many properties of real-world images. The experimental results additionally demonstrate the usability of the generated dataset in real-world applications.

This paper is structured as follows: Section~\ref{sec:previous} gives an overview of the main existing color constancy benchmark datasets, in Section~\ref{sec:crop} the proposed dataset generator is described, in Section~\ref{sec:validation} its properties and capabilities are experimentally validated, and Section~\ref{sec:conclusions} concludes the paper.

\section{Previous work}
\label{sec:previous}

\subsection{Image calibration}
\label{subsec:calibration}

The main idea of color constancy benchmark datasets is for them to have images for which the color of the illumination that influences their scenes is known. That means that along images every such dataset also has the ground-truth illumination for each of these images. For a given image the ground-truth is usually determined by putting a calibration object in the scene and later reading the value of its achromatic surfaces. Calibration objects include gray ball, color checker chart, SpyderCube, etc. Due to the ill-posedness of the illumination estimation problem, determining the ground-truth illumination for a given image without calibration objects can often not be carried out accurately enough. While in such images some of the scene surfaces with known color under the white light could be used, this could lead to inaccuracies due to the metamerism.

\subsection{Existing datasets}
\label{subsec:datasets}

The first large color constancy benchmark dataset with real-world images and ground-truth illumination provided for each image was the GreyBall dataset~\cite{ciurea2003large}. It consists of $11346$ images and in the scene of each image a gray ball is placed and used to determine the ground-truth illumination for this image. However, the images in this dataset are non-linear i.e. they have been processed by applying non-linear operations to them and therefore they do not comply with the image formation model assumed in Eq.~\eqref{eq:image}. Additionally, the images are small with only the of size $240\times 360$.

In 2008 the Color Checker dataset has been proposed~\cite{gehler2008bayesian}. It consists of $568$ images with each of them having a color checker chart in the scene. Several version of the dataset and its ground-truth illumination found their way into the literature over time with most of them being plagued by several serious problems~\cite{finlayson2017curious,hemrit2018rehabilitating,banic2019past}.

Cheng et al. created the NUS dataset in 2014~\cite{cheng2014illuminant}. It is a color constancy dataset composed of natural images captured with 8 different cameras with both indoor and outdoor scenes under various common illuminations. With the same scene taken using multiple cameras, the novelty of this dataset is that the performance of illumination estimation algorithms can be compared across different camera sensors.

In~\cite{banic2017unsupervised} a dataset with 1365 images was published, namely the Cube dataset. It consists of exclusively outdoor images with the SpyderCube calibration object placed in the lower right corner of each image to obtain the ground-truth illumination. All images were taken with the Canon EOS 550D camera. When compared to the previous datasets, the Cube dataset has a higher diversity of scenes and it alleviates some of the previous issues in datasets such as the violation of the uniform illumination assumption. The main disadvantage of the Cube dataset i.e. restriction to only outdoor illuminations was alleviated in the Cube+ dataset~\cite{banic2017unsupervised}. It is a combination of the original Cube dataset and additional 342 images of both indoor scenes and outdoor scenes taken during the night. Consequently, besides the larger number of images, a more diverse distribution of illuminations was achieved which is the desirable property of the color constancy benchmark datasets. All of the newly acquired images in the Cube+ dataset were captured with the same Canon EOS 550D camera and prepared and organized following the same fashion as for the original Cube dataset.

A dataset for camera-independent color constancy was published in~\cite{aytekin2018data}. The images in that dataset were captured with three different cameras with one of them being a mobile phone camera and the other two high-resolution DSLR cameras. The dataset is composed of images in both laboratory and fields scenes taken with all three camera sensors.

Recently a new benchmark dataset with 13k images was introduced~\cite{liu2019self}. It contains both indoor and outdoor scenes with the addition of some challenging images. Unfortunately, at the time, this dataset is not publicly available. Another relatively large dataset with challenging images which is not publicly available was used in~\cite{qiu2018pilot}. Although the authors report the performance of their illumination estimation methods on these datasets, comparison with other methods is hard since they are not publicly available.

During the years of research in the field of color constancy numerous other benchmark datasets such as~\cite{barnard2002comparison1, barnard2002comparison2} were created, but they are not commonly used for the performance evaluation of illumination estimation methods.

\subsection{Problems}
\label{subsec:problems}

The main problem with the previous datasets is the limited number of their images, which is due to the tedious process of the ground-truth illumination extraction. This effectively limits the full-scale application of deep learning methods like for some other problems and various data augmentation techniques have to be used with variable success.

Another problem that can occur during image acquisition is to choose scenes for which the uniform illumination estimation does not hold. This is especially problematic if the less dominant illumination is affecting the calibration object because the extracted ground-truth is then erroneous and results in allegedly hard to estimate image cases~\cite{zakizadeh2015hybrid}.

Even if all of the ground-truth illumination data was correctly collected, it often consists of only the most commonly observed illuminations. This lack of variety makes some of the datasets susceptible to abuse cases of methods that aim to fool some of the error metrics~\cite{banic2015perceptual}. It also prevents the illumination estimation methods from being tested on images formed by the presence of extreme illuminations.

In some of the worst cases, some datasets were used technically inappropriately~\cite{banic2019past}, which made the obtained experimental results to be technically incorrect and put in question some of the allegedly achieved progress~\cite{finlayson2017curious}.

\begin{figure}[htb]
    \centering
    
	\includegraphics[width=\linewidth]{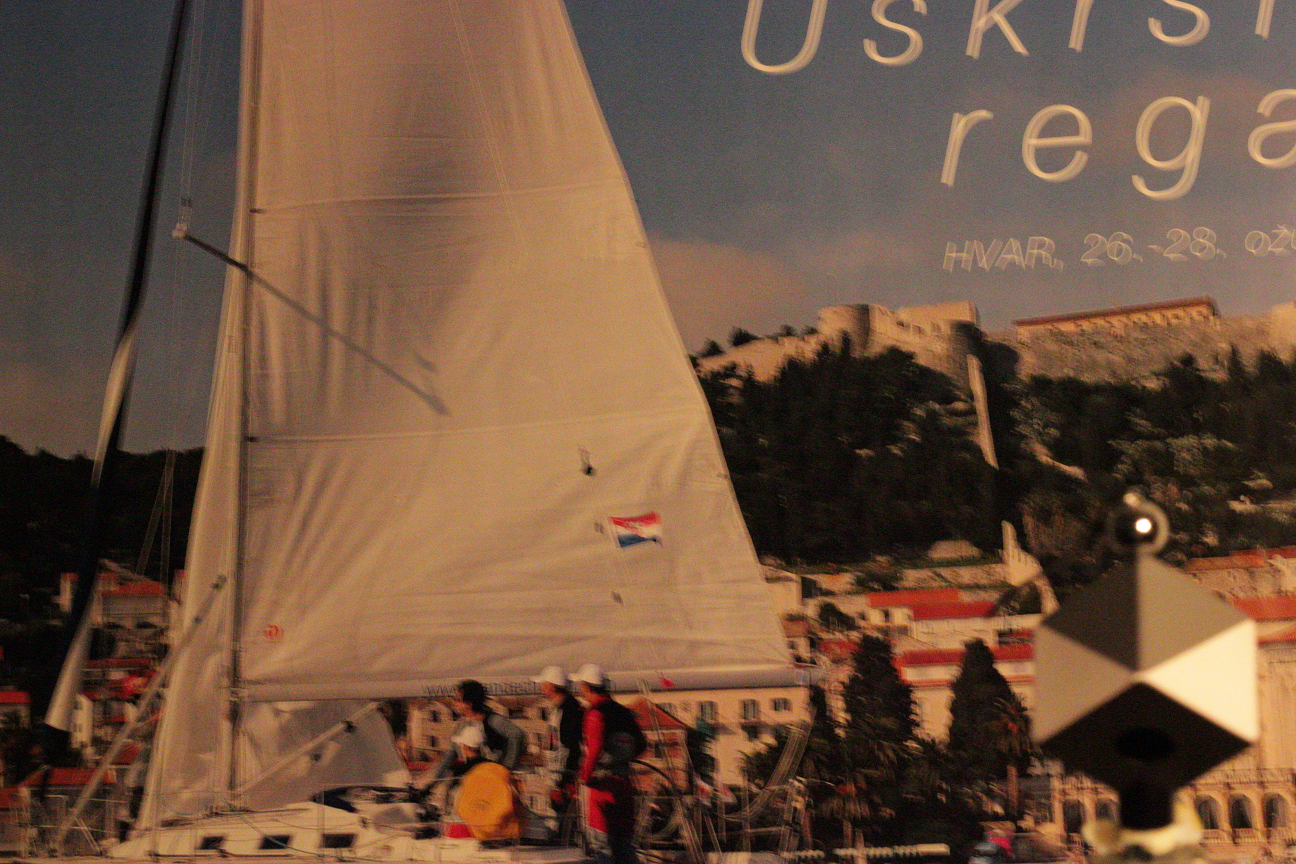}
	
    \caption{Example of an image from the Cube+ dataset~\cite{banic2017unsupervised} whose scene consists only of another printed image.}
	\label{fig:printed}
    
\end{figure}

\section{The proposed dataset generator}
\label{sec:crop}

A solution to many problems mentioned in the previous section would be the possibility to generate real-world images whose scenes are influenced by an arbitrary chosen known illumination and exactly such a solution is proposed in this section. When taking into account everything that has been mentioned here, several conditions have to be met:
\begin{itemize}
\item there has to be a big number of available illuminations,
\item the colors of any material present in the scene that are known for the canonical white illumination have also to be known for every other possible illumination,
\item and the influence of a chosen camera sensor on the color of illuminated material has also to be known.
\end{itemize}

All this can be accomplished by recording enough real-world data and then use it to simulate real-world images. Knowing the behavior of colors of various materials under different illuminations would require too much data both to collect and to control during the image generation process. Because of this and motivated by existence of images like the one in Fig.~\ref{fig:printed}, the proposed dataset generator is restricted only to the colors printed by the same single printer on the same single sheet of paper. To assure uniform illumination and some control over its color, all scenes are illuminated by a projector that projects single color frames. In short, the proposed dataset generator is able to simulate taking of raw camera images of printed images illuminated by a projector. More details are given in the following subsections.

\subsection{Used illuminations}
\label{subsec:illuminations}

\begin{figure}[htb]
    \centering
	\includegraphics[width=\linewidth]{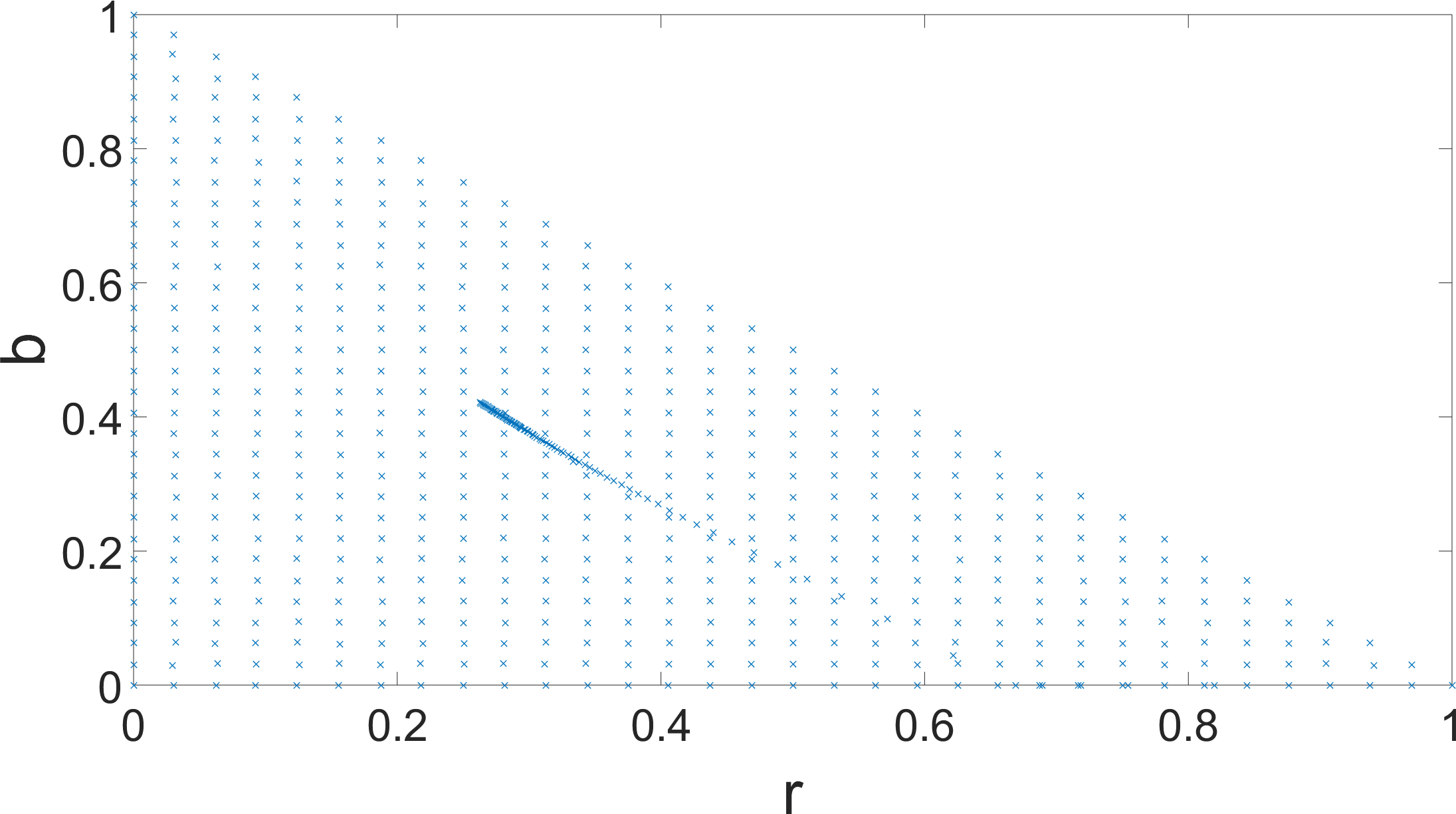}
    \caption{$rb$-chromaticities of the illuminations used to illuminate the printed color pattern.}
	\label{fig:illuminations}
\end{figure}

\begin{figure}[htb]
    \centering
	\includegraphics[width=\linewidth]{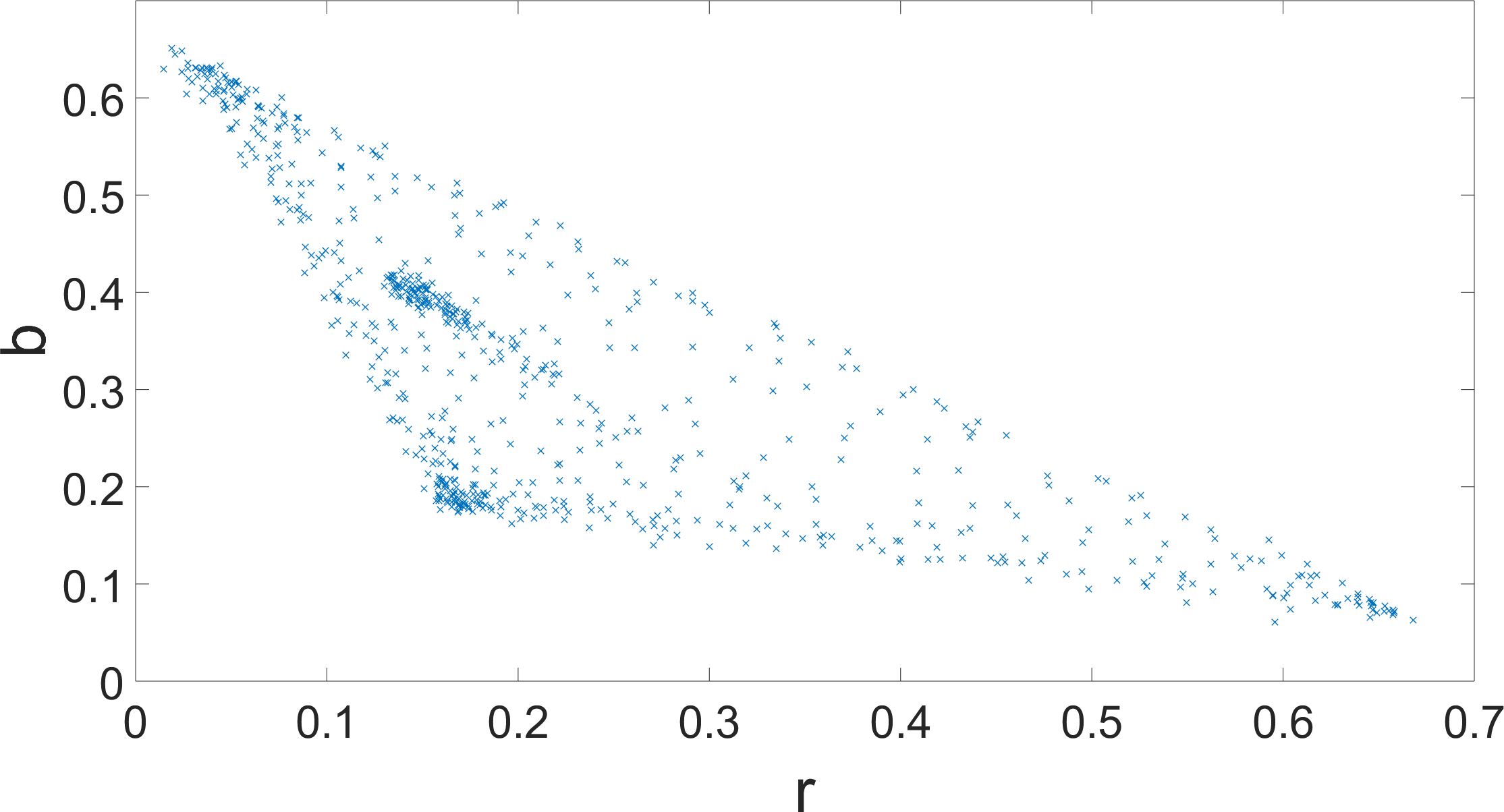}
    \caption{$rb$-chromaticities of the achromatic surfaces of the SpyderCube calibration object colors in the Canon EOS 550D camera RGB after it is illuminated by illuminations with colors from Fig.~\ref{fig:illuminations} and its image taken with a Canon EOS 550D camera.}
	\label{fig:550d}
\end{figure}

\begin{figure}[htb]
    \centering
	\includegraphics[width=\linewidth]{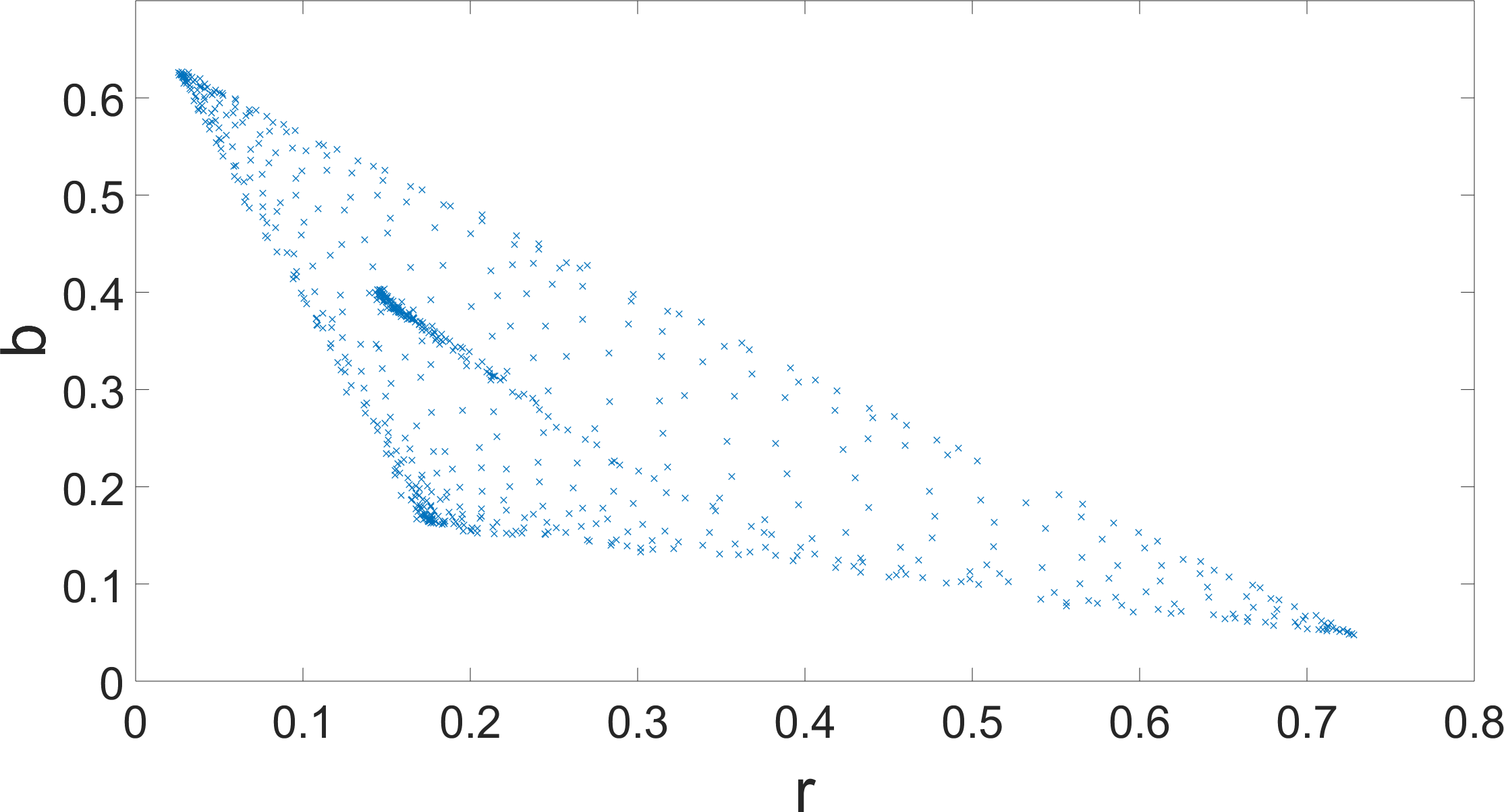}
    \caption{$rb$-chromaticities of the illuminations used to illuminate the printed color pattern.}
	\label{fig:6dmk2}
\end{figure}

To assure a big variability of available illuminations, $707$ of them were used. They are composed of colors whose chromaticities are uniformly spread and of colors of a black body at various temperatures. The latter colors are important because they occur very often in real-world scenes. The relation between all these colors is shown in Fig.~\ref{fig:illuminations}. Due to the projector and camera characteristics, the final appearance of these colors is changed. For example, if the achromatic surfaces of the SpyderCube calibration object are photographed under all these illuminations, their appearances in the RGB colorspaces of two different cameras described in Section~\ref{subsec:cameras} are as shown in Fig.~\ref{fig:550d} and~\ref{fig:6dmk2}.

\begin{figure}[htb]
    \centering
    
	\includegraphics[width=\linewidth]{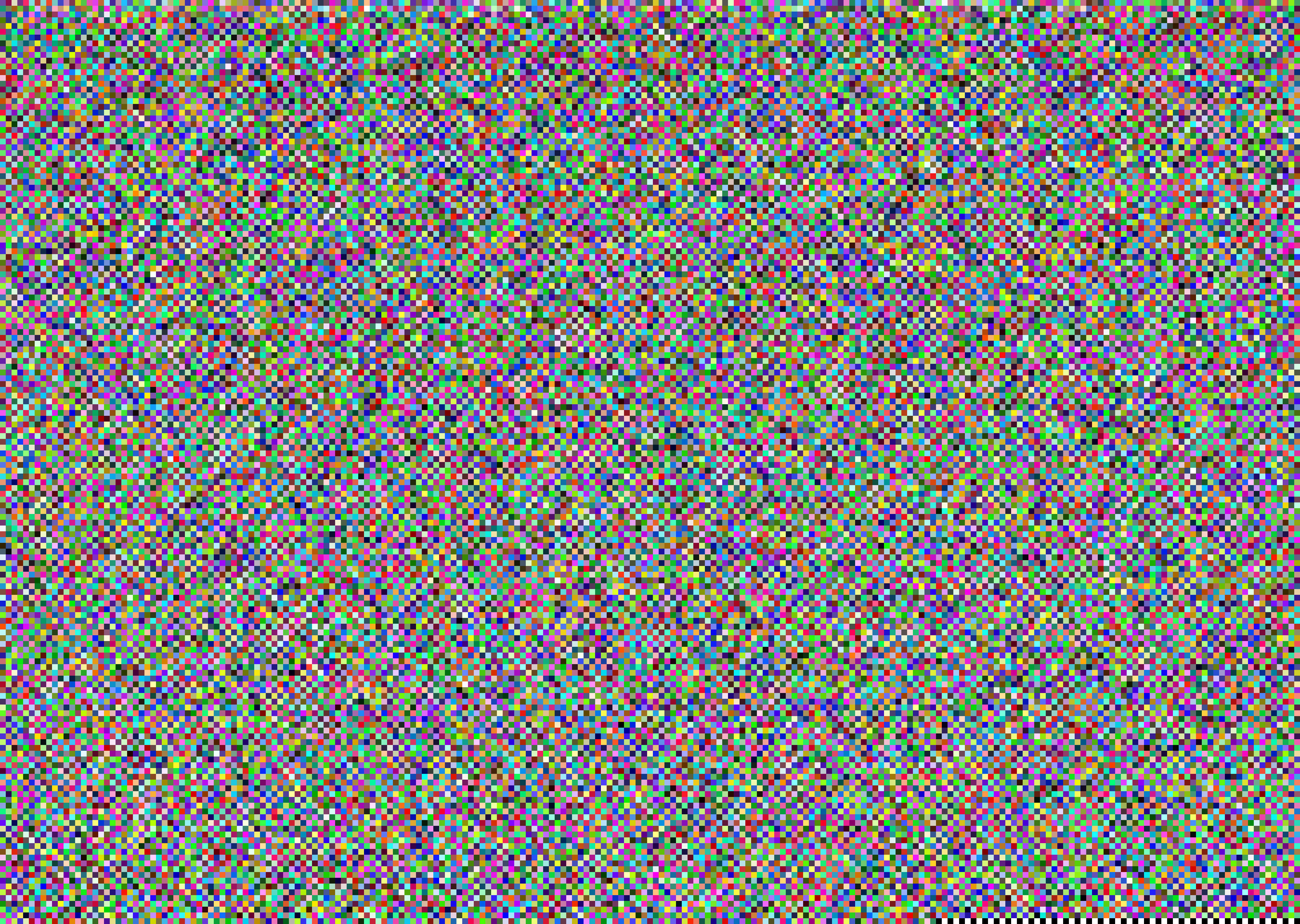}
	
    \caption{Squares in all simplified colors arranged in the pattern that was printed on a single big paper, illuminated by $707$ different illuminations, and photographed.}
	\label{fig:pattern}
    
\end{figure}

\begin{figure*}[htb]
    \centering
    
	\includegraphics[width=\linewidth]{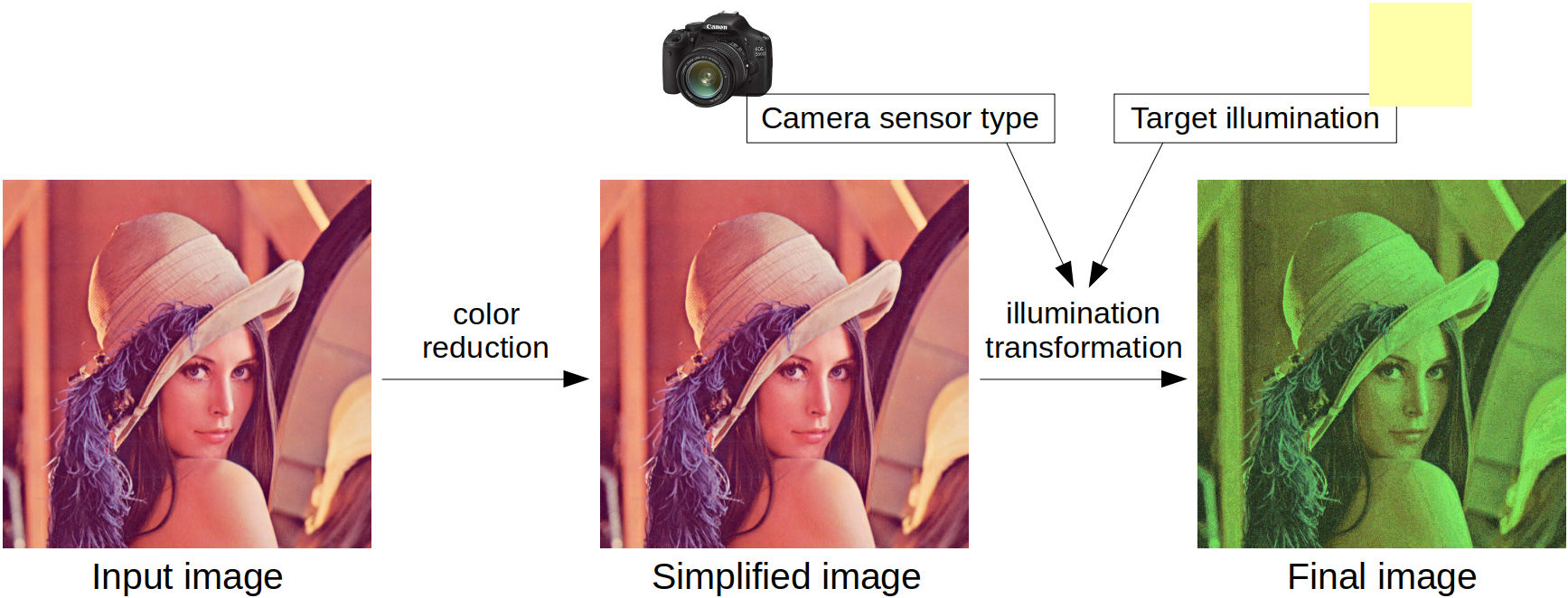}
	
    \caption{The diagram of the image generation process; the Flash tone mapping operator~\cite{banic2016puma, banic2018flash} was used for the final image.}
	\label{fig:generation}
    
\end{figure*}

\subsection{Printed colors}
\label{subsec:printed}
In order to simulate the real-world images, lots of material types would have to be analyzed as the spectral reflectance properties are varying between materials. This is because the material properties determine how a color will change under different illuminations, which is important information for simulating real-world behavior. As handling so much data is hardly feasible in terms of both the data acquisition stage and the image generation stage, the proposed dataset generator uses only one material, namely paper. When printing on paper, RGB colors with 8 bits per channel are used, which leads to a total of $256^3$ i.e. more than 16 million different possible colors. For each of these RBG colors, its behavior when printed on paper has to be known for every illumination chosen in Section~\ref{subsec:illuminations}. Such behavior for a given illumination can be recorded by photographing the printed colors under the projector cast. For the illumination to really be the same for all colors, all of them have to be photographed on the paper simultaneously. Namely, if they were taken partially over several shots, there is the possibility of slight projector cast color changing due to e.g. projector lamp heating. If all $256^3$ colors were used, they could hardly be printed on one paper and later photographed in a high enough resolution. For this reason, instead of using $256^3$ color values, for the proposed generator only $32^3$ were used. They were generated by putting the three least significant bits in the red, green, and blue channel to zero. This number of colors was shown to be appropriate for printing on a single paper sheet of size $A0$, which can be photographed in one shot while still having a high enough resolution. The colors were arranged in the grid shape as shown in Fig.~\ref{fig:pattern}. Each square represents one RBG color under the canonical white illumination. To reflectance properties are constant for each color since they were all printed on the same paper by using the same printer and photographed under the same illumination. Once the printed paper was photographed under all of the $707$ chosen illuminations, a $5\times 5$ pixel area was taken from each of the squares to represent a single color under some illumination. This means that for each of $32^3$ colors there are $25$ realistic representations under for of the $707$ chosen illuminations that can be used to simulate the effects of randomness as well as noise.

\subsection{Generator cameras}
\label{subsec:cameras}
The printed color pattern was photographed under different illuminations with two Canon cameras, namely Canon EOS 550D and Canon EOS 6D Mark II. In order to obtain the linear PNG images that comply with the model in Eq.~\eqref{eq:image} from raw images, the \texttt{dcraw} tool with options \texttt{-D -4 -T} was used followed by simple subsampling and debayering. The sensor field resolution for the former Canon camera is $5202\times 3465$, whereas the latter camera model has the sensor field resolution of $6384\times 4224$. Higher camera resolution enables higher precision when extracting the color values from the squares of the photographed color pattern as the boundaries of squares tend to get blurred when using lower resolution images. By comparing Fig.~\ref{fig:550d} and~\ref{fig:6dmk2}, which show the $rb$-chromaticities of the illuminations captured with two cameras, the difference in $rb$-chromaticities of the illuminations can be noticed. This clearly shows how camera sensor characteristics differ, with the Canon EOS 6D Mark II producing smoother illumination estimations.

\subsection{Image generation}
\label{subsec:generation}
Generating a new image includes choosing the source image, the desired illumination, and the camera sensor. The source image is first simplified following the same procedure as for the creation of the color pattern described in Section~\ref{subsec:printed}, i.e. the three least significant bits in the red, green, and blue channel are put to zero. That way, the colors in the source image are constrained to the ones in the color pattern shown in Fig.~\ref{fig:pattern}, whose behavior on paper under the previously selected illumination is known. Then, the color of every pixel in the simplified image is changed to a color observed on the pattern square of the same color when it was photographed under the desired illumination. As mentioned earlier, there are $25$ possible choices for this change. Doing this for all pixels gives a raw linear image as if the initially chosen image is printed, illuminated by the projector using the initially chosen illumination, and then photographed. Fig.~\ref{fig:generation} illustrates the described steps for the whole image generation process. Repeating this procedure by having a fixed camera sensor results in a new dataset.

\begin{figure}[htb]
    \centering
    
	\includegraphics[width=\linewidth]{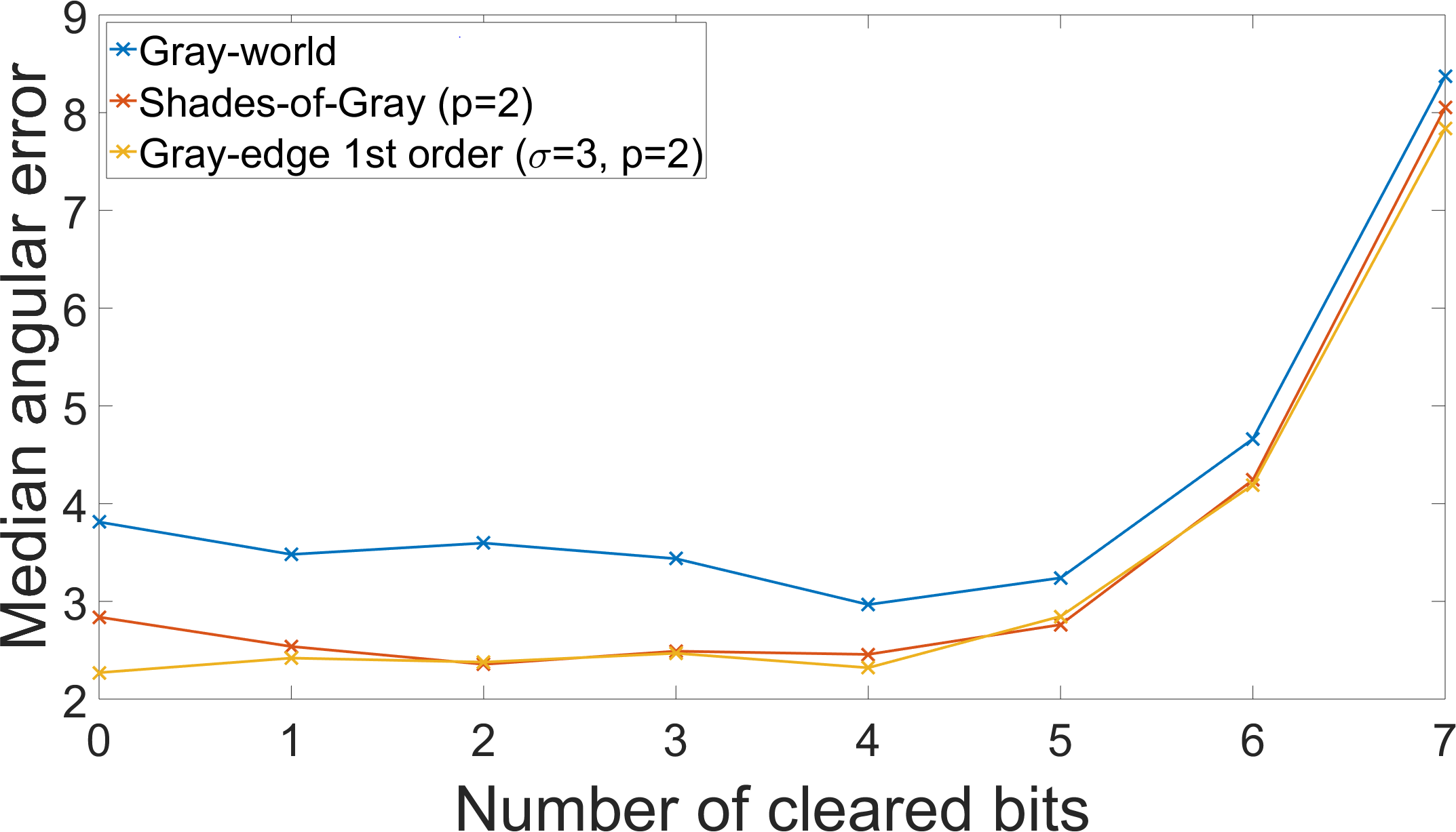}
	
    \caption{The effect of color reduction on the performance of illumination estimation methods.}
	\label{fig:bits}
    
\end{figure}
\begin{figure*}[htb]
    \centering
  
  \subfloat[]{
  \includegraphics[width=0.115\linewidth]{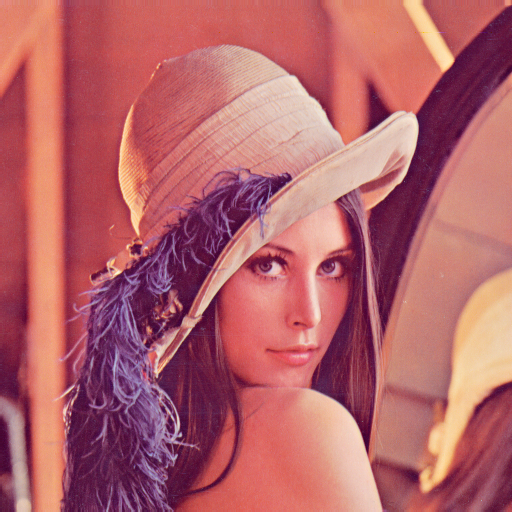}
  \label{fig:lena_0}
  }%
  \subfloat[]{
  \includegraphics[width=0.115\linewidth]{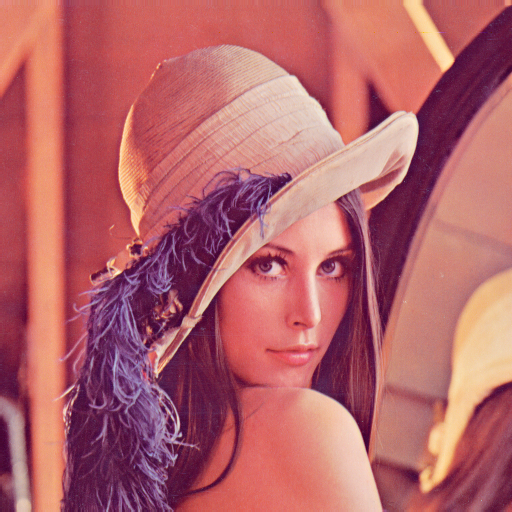}
  \label{fig:lena_1}
  }%
  \subfloat[]{
  \includegraphics[width=0.115\linewidth]{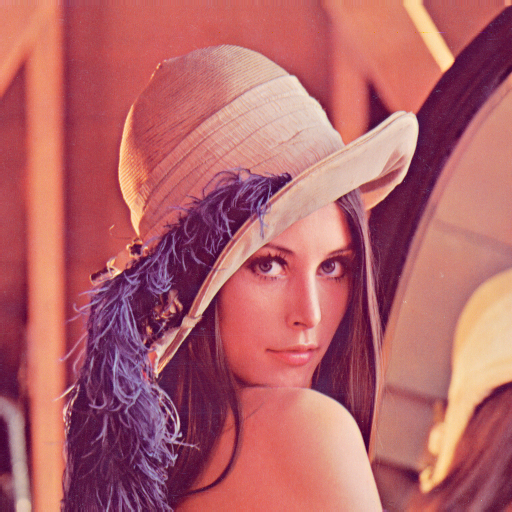}
  \label{fig:lena_2}
  }%
  \subfloat[]{
  \includegraphics[width=0.115\linewidth]{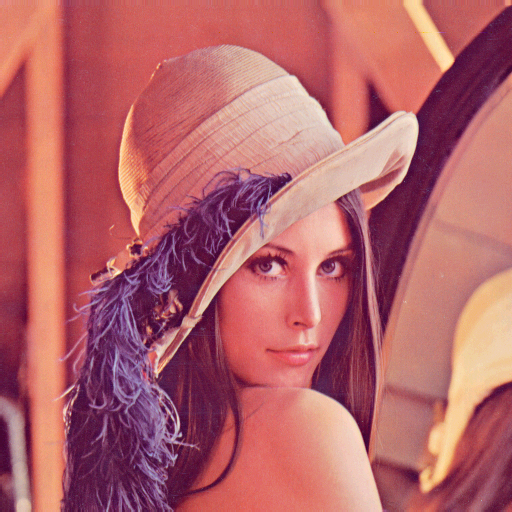}
  \label{fig:lena_3}
  }%
  \subfloat[]{
  \includegraphics[width=0.115\linewidth]{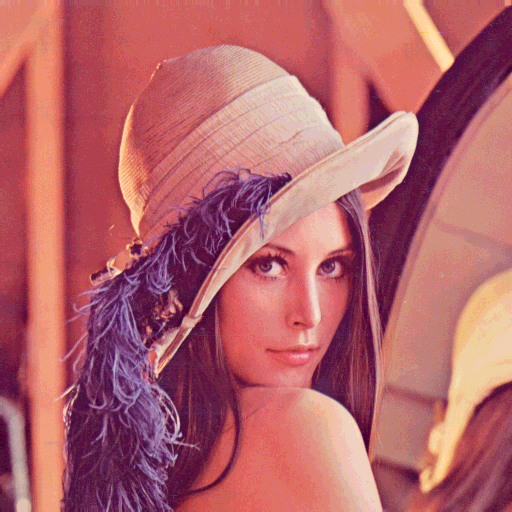}
  \label{fig:lena_4}
  }
  \subfloat[]{
  \includegraphics[width=0.115\linewidth]{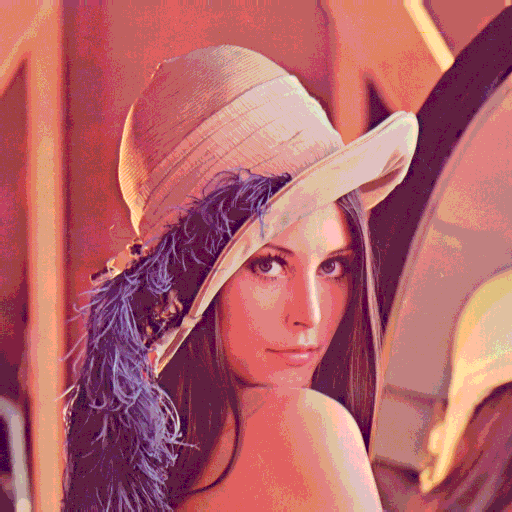}
  \label{fig:lena_5}
  }
  \subfloat[]{
  \includegraphics[width=0.115\linewidth]{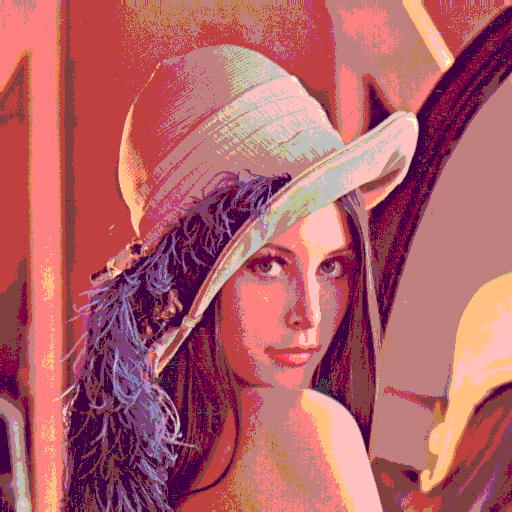}
  \label{fig:lena_6}
  }
  \subfloat[]{
  \includegraphics[width=0.115\linewidth]{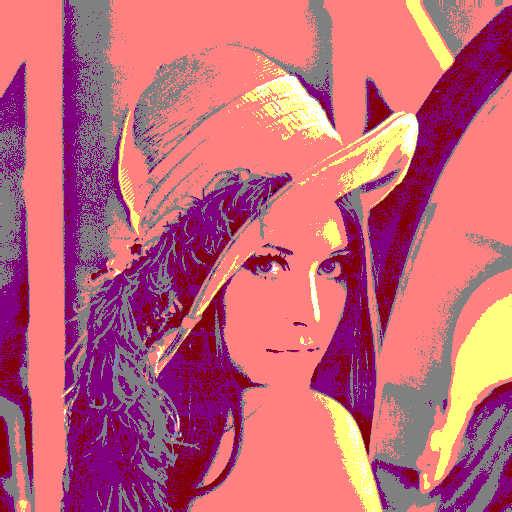}
  \label{fig:lena_8}
  }
  
    \caption{Influence of color reduction: (a)~without color reduction; (b) to (h)~with color reduction, starting with only one bit i the red, green, and blue channel put to zero for (b) up to  seven bits for (h).}
  \label{fig:reduction}
    
\end{figure*}

\subsection{Name}
\label{subsec:name}

Since the color pattern used to create the proposed dataset generator was printed in Croatia and all scenes were illuminated and photographed in Croatia, the proposed dataset generator was simply named Croatian Paper~(CroP).

\section{Experimental validation}
\label{sec:validation}

\subsection{Error metrics}
\label{subsec:metrics}
The angular error is the most commonly used among many error metrics that have been proposed to measure the performance of illumination estimations methods~\cite{gijsenij2009perceptual,banic2015perceptual}. There are two kinds of angular error, namely the recovery angular error and the reproduction angular error. When neither of these two is explicitly mentioned, it is commonly understood that the recovery angular error is used. The recovery angular error is defined as the angle between the illumination estimation and the ground-truth illumination
\begin{equation}
\label{eq:recovery}
    err_{recovery} = cos^{-1}\bigg(\frac{\boldsymbol{\rho}^E\cdot\boldsymbol{\rho}^{Est}}{\norm{\boldsymbol{\rho}^E}\norm{\boldsymbol{\rho}^{Est}}}\bigg)
\end{equation}
where the $\boldsymbol{\rho}^{Est}$ is the illumination estimation, $\boldsymbol{\rho}^E$ is the ground-truth illumination, and '$\cdot$' is the vector dot product. The reproduction angular error~\cite{finlayson2014reproduction,finlayson2017reproduction} has been defined as
\begin{equation}
\label{eq:reproduction}
    err_{reproduction} = cos^{-1}\Bigg(\frac{(\boldsymbol{\rho}^{E,W}/\boldsymbol{\rho}^{Est})\cdot\boldsymbol{U}}{\abs{\boldsymbol{\rho}^{E, W}/\boldsymbol{\rho}^{Est}}\sqrt{3}}\Bigg)
\end{equation}
where $\boldsymbol{\rho}^{E, W}$ is the vector of the white surface color in the image RGB color space under the scene illumination, $\mathbf{U}$ is the vector of the ideally corrected white color, i.e. $[1,1,1]^T$. Although the recovery angular error has been and still is extensively used, it has been shown in~\cite{finlayson2017reproduction} how the change in the illumination of the same scene can cause significant fluctuations of the recovery angular error, while the reproduction angular error has been shown to be stable.

To evaluate the illumination estimation method performance on a whole dataset, the error values calculated for all dataset images are summed up using various summary statistics. As the distribution of the angular errors is non-symmetrical, it is much better to use the median instead of the mean angular error~\cite{hordley2004re}. However, other measures such as mean, trimean, and best and worst $p\%$ are also used for additional comparisons of methods. In~\cite{barron2015convolutional} the measure often called as the average was introduced. It is the geometric mean of the mean, median, trimean, best $25\%$, and worst $25\%$ of the obtained angular errors. In the following experiments, the median angular error of the reproduction angular error has been used as the reference summary statistic.

\subsection{Influence of color reduction}
\label{subsec:reduction}
As described in Sections~\ref{subsec:printed} and~\ref{subsec:generation}, the number of colors in both the printed pattern and the input image are reduced to the total of $32^3$ different colors by setting the three least significant bits in the red, green, and blue channel to zero. Fig.~\ref{fig:reduction} shows how this type of color reduction influences the quality of sRGB images for different number of bits being set to zero. To test the effect of bits removal on the performance of illumination estimation methods, linear images of the Canon 1Ds Mk III dataset from the NUS datasets~\cite{cheng2014illuminant} were used. Since the dataset generator manages bits on sRGB images, for the sake of simulating bits removal the linear images were first tone mapped and converted to sRGB images with 8 bits per channel by applying the Flash tone mapping operator~\cite{banic2016puma,banic2018flash}. Next, the three least significant bits were set to zero, and then the image was returned to its linear form by applying the reversed formula of the Flash tone mapping operator. Finally, illumination estimation methods were applied to such changed images. The results for Gray-world~\cite{buchsbaum1980spatial}, Shades-of-Gray~\cite{finlayson2004shades}, and \nth{1} order Gray-Edge~\cite{van2007edge} applied on raw images with reduced colors are shown in Fig.~\ref{fig:bits}. In some cases of bits clearing the median angular error for Gray-World and Shades-of-Gray methods is better than when the original linear images are used. Since bits clearing can eliminate darker pixels, this reminds of~\cite{joze2012role} where using only bright pixels for illumination estimation resulted in improved accuracy. As opposed to that, the \nth{1} order Gray-Edge method did not improve when removing the bits. This method relies on the edge information to estimate the illuminations and in that case the color reduction can be detrimental since it can reduce edges.

\begin{table}[ht]
\scriptsize
\caption{Performance of White-Patch~\cite{funt2010rehabilitation}, Gray-world~\cite{buchsbaum1980spatial}, and Shades-of-Gray~\cite{finlayson2004shades} on 8 generated datasets (lower Avg. is better). The used format is the same as in~\cite{barron2015convolutional}. ''C1'' is the abbreviation for Canon 1Ds Mk III dataset, which is one of NUS datasets~\cite{cheng2014illuminant}, ''550D'' represents Canon EOS 550D camera, and ''6D'' represents Canon 6D Mark II camera.}
\label{tab:performance}
\centering
\begin{tabular}{|>{\centering}m{2.75cm}| >{\centering}m{3mm} >{\centering}m{3mm} >{\centering\arraybackslash}m{3mm} >{\centering\arraybackslash}m{3mm} >{\centering\arraybackslash}m{3mm} >{\centering\arraybackslash}m{3mm}|}

    \hline
    \multicolumn{7}{|c|}{\makecell{C1 scenes,~~~~~6D sensor,~~~~~C1 illuminations}}\\
    \hline
    \textbf{Algorithm} & \textbf{Mean} & \textbf{Med.} & \textbf{Tri.} & \makecell{\textbf{Best}\\\textbf{25\%}} & \makecell{\textbf{Worst}\\\textbf{25\%}} & \textbf{Avg.}\\
    
    \hline
    White-Patch~\cite{funt2010rehabilitation} & 2.61 & 2.59 & 2.50 & 1.03 & 4.38 & 2.38\\
    Gray-world~\cite{buchsbaum1980spatial} & 6.27 & 5.32 & 5.58 & 3.34 & 10.75 & 5.82\\
    Shades-of-Gray (p=2)~\cite{finlayson2004shades} & 2.79 & 2.36 & 2.40 & 1.28 & 5.12 & 2.53\\
    
    \hline
    \hline
    \multicolumn{7}{|c|}{\makecell{C1 scenes,~~~~~6D sensor,~~~~~Random illuminations}}\\
    \hline
    \textbf{Algorithm} & \textbf{Mean} & \textbf{Med.} & \textbf{Tri.} & \makecell{\textbf{Best}\\\textbf{25\%}} & \makecell{\textbf{Worst}\\\textbf{25\%}} & \textbf{Avg.}\\
    
    \hline
    White-Patch~\cite{funt2010rehabilitation} & 2.17 & 2.05 & 2.08 & 0.88 & 3.73 & 1.98\\
    Gray-world~\cite{buchsbaum1980spatial} & 5.79 & 5.20 & 5.38 & 2.64 & 9.82 & 5.30\\
    Shades-of-Gray (p=2)~\cite{finlayson2004shades} & 2.34 & 1.93 & 1.96 & 0.98 & 4.43 & 2.08\\
    
    \hline
    \hline
    \multicolumn{7}{|c|}{\makecell{C1 scenes,~~~~~550D sensor,~~~~~C1 illuminations}}\\
    \hline
    \textbf{Algorithm} & \textbf{Mean} & \textbf{Med.} & \textbf{Tri.} & \makecell{\textbf{Best}\\\textbf{25\%}} & \makecell{\textbf{Worst}\\\textbf{25\%}} & \textbf{Avg.}\\
    
    \hline
    White-Patch~\cite{funt2010rehabilitation} & 9.41 & 5.38 & 5.56 & 2.60 & 23.56 & 7.04\\
    Gray-world~\cite{buchsbaum1980spatial} & 5.75 & 5.25 & 5.39 & 2.75 & 9.45 & 5.31\\
    Shades-of-Gray (p=2)~\cite{finlayson2004shades} & 2.61 & 2.07 & 2.14 & 0.97 & 5.20 & 2.25\\
    
    \hline
    \hline
    \multicolumn{7}{|c|}{\makecell{C1 scenes,~~~~~550D sensor,~~~~~Random illuminations}}\\
    \hline
    \textbf{Algorithm} & \textbf{Mean} & \textbf{Med.} & \textbf{Tri.} & \makecell{\textbf{Best}\\\textbf{25\%}} & \makecell{\textbf{Worst}\\\textbf{25\%}} & \textbf{Avg.}\\
    
    \hline
    White-Patch~\cite{funt2010rehabilitation} & 10.90 & 6.75 & 6.81 & 2.90 & 27.18 & 8.31\\
    Gray-world~\cite{buchsbaum1980spatial} & 5.25 & 5.04 & 5.07 & 2.65 & 8.32 & 4.94\\
    Shades-of-Gray (p=2)~\cite{finlayson2004shades} & 2.15 & 1.73 & 1.83 & 0.67 & 4.35 & 1.82\\
    
    \hline
    \hline
    \multicolumn{7}{|c|}{\makecell{Random scenes,~~~~~6D sensor,~~~~~C1 illuminations}}\\
    \hline
    \textbf{Algorithm} & \textbf{Mean} & \textbf{Med.} & \textbf{Tri.} & \makecell{\textbf{Best}\\\textbf{25\%}} & \makecell{\textbf{Worst}\\\textbf{25\%}} & \textbf{Avg.}\\
    
    \hline
    White-Patch~\cite{funt2010rehabilitation} & 2.59 & 2.23 & 2.37 & 1.31 & 4.31 & 2.38\\
    Gray-world~\cite{buchsbaum1980spatial} & 3.84 & 4.06 & 3.96 & 3.06 & 4.34 & 3.82\\
    Shades-of-Gray (p=2)~\cite{finlayson2004shades} & 2.73 & 2.78 & 2.78 & 1.95 & 3.22 & 2.66\\
    
    \hline
    \hline
    \multicolumn{7}{|c|}{\makecell{Random scenes,~~~~~6D sensor,~~~~~Random illuminations}}\\
    \hline
    \textbf{Algorithm} & \textbf{Mean} & \textbf{Med.} & \textbf{Tri.} & \makecell{\textbf{Best}\\\textbf{25\%}} & \makecell{\textbf{Worst}\\\textbf{25\%}} & \textbf{Avg.}\\
    
    \hline
    White-Patch~\cite{funt2010rehabilitation} & 2.46 & 2.15 & 2.33 & 0.88 & 4.34 & 2.16\\
    Gray-world~\cite{buchsbaum1980spatial} & 4.09 & 4.16 & 4.20 & 2.53 & 5.38 & 3.96\\
    Shades-of-Gray (p=2)~\cite{finlayson2004shades} & 2.47 & 2.64 & 2.57 & 1.54 & 3.17 & 2.42\\
    
    \hline
    \hline
    \multicolumn{7}{|c|}{\makecell{Random scenes,~~~~~550D sensor,~~~~~C1 illuminations}}\\
    \hline
    \textbf{Algorithm} & \textbf{Mean} & \textbf{Med.} & \textbf{Tri.} & \makecell{\textbf{Best}\\\textbf{25\%}} & \makecell{\textbf{Worst}\\\textbf{25\%}} & \textbf{Avg.}\\
    
    \hline
    White-Patch~\cite{funt2010rehabilitation} & 22.79 & 10.35 & 19.52 & 6.43 & 51.43 & 17.24\\
    Gray-world~\cite{buchsbaum1980spatial} & 3.99 & 4.28 & 4.14 & 2.16 & 5.65 & 3.86\\
    Shades-of-Gray (p=2)~\cite{finlayson2004shades} & 2.36 & 2.43 & 2.31 & 1.13 & 3.68 & 2.23\\
    
    \hline
    \hline
    \multicolumn{7}{|c|}{\makecell{Random scenes,~~~~~550D sensor,~~~~~Random illuminations}}\\
    \hline
    \textbf{Algorithm} & \textbf{Mean} & \textbf{Med.} & \textbf{Tri.} & \makecell{\textbf{Best}\\\textbf{25\%}} & \makecell{\textbf{Worst}\\\textbf{25\%}} & \textbf{Avg.}\\
    
    \hline
    White-Patch~\cite{funt2010rehabilitation} & 25.81 & 12.30 & 21.33 & 7.45 & 59.80 & 19.77\\
    Gray-world~\cite{buchsbaum1980spatial} & 4.25 & 4.23 & 4.20 & 2.44 & 6.15 & 4.08\\
    Shades-of-Gray (p=2)~\cite{finlayson2004shades} & 4.01 & 2.80 & 2.81 & 0.85 & 9.69 & 3.04\\
    \hline

\end{tabular}
\end{table}

\subsection{Method performance}
\label{subsec:performace}
Several dataset were created to evaluate the behavior of some simpler illumination estimation methods on generated images and compare it to the behavior on real-world datasets. To create the test datasets, two options were used for the scenes whose printing was to be simulated, two options were used for the camera sensors, and two options were used for the illuminations. When these options were combined through Cartesian product, they resulted in $8$ triplets of inputs for the proposed dataset generator and consequently in $8$ datasets. Two options for the scenes were the sRGB images of the Canon 1Ds Mk III dataset, which is one of the NUS datasets~\cite{cheng2014illuminant}, and synthetic images where all pixel values were randomly drawn from uniform distribution. The camera options included Canon EOS 550D and Canon 6D Mark II. As for the illuminations, the mentioned two options were a subset of illuminations from Section~\ref{subsec:illuminations} that are closest to the ground-truth illuminations of Canon 1Ds Mk III dataset and a subset of randomly chosen illuminations described in Section~\ref{subsec:illuminations}. The results for White-Patch~\cite{funt2010rehabilitation}, Gray-world~\cite{buchsbaum1980spatial}, and Shades-of-Gray~\cite{finlayson2004shades} on the $8$ generated datasets are reported in Table~\ref{tab:performance}. The obtained angular error statistics and their relations for different methods are very similar to the ones obtained on other well known real-world datasets~\cite{cheng2014illuminant,banic2017unsupervised}. Particularly interesting are the results of the White-patch method. Namely, for the datasets where the Canon~EOS~6D~Mk~II camera was used, the White-patch method performed surprisingly well when compared to the datasets where the Canon~EOS~550D camera was used. This can be attributed to higher resolution of the former Canon camera as well as of its higher sensor quality due to its being of a significantly newer production date. In other words, the datasets where the Canon~EOS~550D camera was used contain more noise then the ones where for the Canon~EOS~6D~Mk~II camera.


\begin{table}[ht]
\scriptsize
\caption{Comparison of performance of some learning-based methods on the Cube+ dataset~\cite{banic2017unsupervised} with respect to the training (lower Avg. is better). The used format is the same as in~\cite{barron2015convolutional}.}
\label{tab:cube_plus}
\centering
\begin{tabular}{|>{\centering}m{3.9cm}| >{\centering}m{3mm} >{\centering}m{3mm} >{\centering\arraybackslash}m{3mm} >{\centering\arraybackslash}m{3mm} >{\centering\arraybackslash}m{3mm} >{\centering\arraybackslash}m{3mm}|}

    \hline
    \textbf{Algorithm} & \textbf{Mean} & \textbf{Med.} & \textbf{Tri.} & \makecell{\textbf{Best}\\\textbf{25\%}} & \makecell{\textbf{Worst}\\\textbf{25\%}} & \textbf{Avg.}\\
    
    \hline
    
    
    \multicolumn{7}{|c|}{Trained and tested Cube+ dataset (through cross-validation)}\\
    \hline
    Smart Color Cat~\cite{banic2015using} & 2.27 & 1.35 & 1.61 & 0.34 & 5.72 & 1.58\\
	Regression trees (simple features)~\cite{cheng2015effective} & 1.57 & 0.89 & 1.04 & 0.20 & 4.15 & 1.04\\
	Color Beaver (using Gray-world)~\cite{koscevic2019color} & 1.49 & 0.77 & 0.98 & 0.21 & 3.94 & 0.99\\
    \hline
    \multicolumn{7}{|c|}{Trained on the generated dataset and tested on the Cube+ dataset}\\
    \hline
    Regression trees (simple features)~\cite{cheng2015effective} & 2.54 & 1.66 & 1.89 & 0.45 & 6.07 & 1.85\\
	Smart Color Cat~\cite{banic2015using} & 2.47 & 1.43 & 1.76 & 0.40 & 6.21 & 1.73\\
	Color Beaver (using Gray-world)~\cite{koscevic2019color} & 1.73 & 0.74 & 0.97 & 0.37 & 4.75 & 1.17\\
	
    \hline
    
\end{tabular}
\end{table}

\subsection{Real-world performance}
\label{subsec:real}

To check to what degree the datasets generated by the proposed dataset generator resemble the real-world and help coping with it, an experiment with the Cube+ dataset~\cite{banic2017unsupervised} was carried out. This dataset happens to consist of images taken by the very same Canon~EOS~550D camera the was used during the creation of the proposed dataset generator. Therefore, the proposed dataset generator was used to simulate the use of the Canon~EOS~550D camera to take photos of printed sRGB Cube+ images illuminated by the illuminations similar to  Cube+ ground-truth illuminations.

Several learning-based methods were then first trained on the artificially generated dataset and tested on the real-world Cube+ dataset. The obtained results are shown in Table~\ref{tab:cube_plus}. Training on real-world images is obviously better, but for methods like Color Beaver the difference in performance with respect to the used training data is not too big and statistics like the median and the trimean angular error are even better. For the Smart Color Cat method the number of bins was restricted due to the colors themselves being restricted. As for the regression trees, their performance was affected the most, but they still obtained relatively accurate results. Some of the performance degrading may be attributed to the Canon~EOS~550D data having more noise as previously mentioned, while for Canon~EOS~6D~Mk~II a similar experiment could not have been conducted since it was not used to create any real-world public dataset.

The obtained results can be said to serve as a proof-of-concept that learning from realistically generated artificial images can lead to high accuracy on the real-world images.

\subsection{Comparison to datasets with real-world images}
\label{subsec:comparison}

Some of the advantages of using the proposed CroP are:
\begin{itemize}
    \item there is a large variety of possible illuminations that can be used when images are being created and the illumination distribution can easily be controlled
    \item the images contain no calibration objects that would have to be masked out to prevent any unfair bias,
    \item there is no black level and there are no clipped pixels,
    \item the generated images can be influenced by arbitrary many illuminations with clearly defined ground-truth,
    \item the number of dataset images can be arbitrarly high.
\end{itemize}

Some of the disadvantages of the proposed CroP include:
\begin{itemize}
    \item only one material i.e. paper is used in all images,
    \item the spectral characteristics of the illuminations are limited by the ones of the lamps in the used projector.
\end{itemize}

\section{Conclusions}
\label{sec:conclusions}
In this paper, a color constancy dataset generator that enables generating realistic linear raw images has been proposed. While image generation is constrained to a smaller subset of possible realistic images, these have been shown to share many properties with the real-world images when statistics-based methods are applied to them. Additionally, it has been demonstrated that these images can be used to train learning-based methods, which then achieve relatively accurate results on the real-world datasets. This potentially means that the proposed dataset generator could be used to create large amounts of images required for some more advanced deep learning techniques. Future work will include experiments with generating images with multiple illuminations and adding new camera models and illuminations.

\section*{Acknowledgment}

This work has been supported by the Croatian Science Foundation under Project IP-06-2016-2092.

{\small
\balance
\bibliographystyle{ieee}
\bibliography{crop}
}

\end{document}